\documentclass[11pt]{article}

\usepackage[T1]{fontenc}
\usepackage[utf8]{inputenc}
\usepackage{lmodern}

\usepackage{amsmath, amssymb, amsthm}

\usepackage{microtype}
\usepackage{geometry}
\usepackage{graphicx}
\usepackage{booktabs}

\usepackage[hidelinks]{hyperref}

\usepackage[numbers]{natbib}

\usepackage{authblk}

\theoremstyle{plain}

\theoremstyle{definition}

\usepackage{tabularx}
\usepackage{array}
\usepackage{booktabs}

\usepackage[font=small]{caption}
\usepackage{subcaption}

\usepackage{float}

\usepackage[section]{placeins}

\newcolumntype{L}{>{\raggedright\arraybackslash}X}

\title{From Imitation to Interaction: Mastering Game of Schnapsen with Shallow Reinforcement Learning}

\author[1]{Ján Klačan}
\author[2]{Sizhong Zhang}

\affil[1]{Vrije Universiteit Amsterdam \\\texttt{j.klacan@student.vu.nl}}
\affil[2]{Vrije Universiteit Amsterdam \\\texttt{s.zhang9@student.vu.nl}}

\date{\today}

\begin{document}

\setcounter{topnumber}{2}
\setcounter{bottomnumber}{2}
\setcounter{totalnumber}{4}
\renewcommand{\topfraction}{0.85}
\renewcommand{\bottomfraction}{0.85}
\renewcommand{\textfraction}{0.10}
\renewcommand{\floatpagefraction}{0.75}

\maketitle

\begin{abstract}
This paper investigates whether shallow neural network agents can master the card game Schnapsen and challenge a strong search-based baseline, RdeepBot, which uses Monte Carlo sampling and lookahead search. Guided by a progressively more complex experimental design, we first evaluate a supervised learning agent (MLPBot) trained on replay data and then a reinforcement learning agent (RLBot) with the same shallow architecture trained through asynchronous Monte Carlo updates and experience replay. The results show that supervised imitation does not generalize well enough to defeat strong RdeepBot opponents, whereas reinforcement learning produces substantially stronger agents. In the setting that focuses on the \texttt{depth} parameter of RdeepBot, the best performance is achieved when the learned value function is combined with deeper lookahead during gameplay, allowing RLBot to achieve statistically significant higher winning rates against the strongest evaluated RdeepBot baseline. In the sample-based setting, the gains are more conditional: the strongest performance appears at a relatively lower training \texttt{num\_samples} parameter rather than increasing uniformly with stronger sampling.

\end{abstract}


\section{Introduction}

Schnapsen is a popular trick-taking card game that presents unique challenges for artificial intelligence due to its hidden information, changing game phases, and complex strategic state space. Various bots have been developed to master the game, ranging from randomized algorithms to sophisticated rule-based agents. A strong example of the latter is RdeepBot, an agent included in the Schnapsen Game Engine \cite{isc_schnapsen_repo}. RdeepBot handles imperfect information through Monte Carlo sampling and evaluates actions through fixed lookahead search, which makes it a competitive but computationally expensive option.

In contrast, neural-network-based agents do not rely on explicitly encoded strategies - instead, they must learn to approximate the value of game situations from data or gameplay interaction. In this paper, we investigate whether shallow neural networks can learn strategies that are strong enough to outperform RdeepBot. Following a progressive experimental design, we begin with the simplest learning setup and increase complexity only when necessary.

First, we study a custom supervised learning agent inspired by the machine learning bot in the Schnapsen Game Engine \cite{isc_schnapsen_repo}. We call our bot the MLPBot, and it is trained on a static replay dataset generated from games played by RdeepBot variants. This experiment tests whether imitation of a search-based teacher is sufficient for robust play in Schnapsen. Second, we develop a reinforcement learning agent, RLBot, with the same shallow network architecture but a different training procedure based on asynchronous Monte Carlo updates and experience replay \cite{lin1992self}. Instead of reproducing decisions from a fixed dataset, RLBot learns from gameplay interaction and is evaluated against RdeepBot variants with different search depths and \texttt{num\_samples} settings.

The results show a clear difference between the two learning paradigms. Static imitation does not generalize well enough to defeat search-based opponents, whereas shallow reinforcement learning produces substantially stronger agents. In a setting focused on search \texttt{depth} of RdeepBot, the best results are obtained when the learned value function is combined with deeper lookahead during gameplay. In the sample-based setting focused on the \texttt{num\_samples} parameter, reinforcement learning also yields strong agents. However, the effect is more conditional: the best performance is obtained only for a limited range of training \texttt{num\_samples} values rather than increasing uniformly with stronger sampling.

\section{Background Information}

\subsection{Rules of Card Game Schnapsen}
This research utilizes Schnapsen, a two-player trick-taking game with 20 cards \cite{psellos_schnapsen}. A deal is won by reaching 66 trick points or by taking the last trick. Points result from capturing cards (Ace=11, Ten=10, King=4, Queen=3, Jack=2) and declaring marriages (King-Queen pairs: 40 points for trump, 20 otherwise). Players begin with five cards, and a face-up card is revealed that determines the trump suit. Tricks are won by the highest trump or highest card of the led suit, and the winner leads the next trick. Gameplay is divided into two phases. In Phase 1 (when talon is open), players draw replacements, are not required to follow suit, and can exchange the lowest trump for the face-up one. The information is imperfect. Phase 2 (talon exhausted or voluntarily closed) has stricter rules: players must follow suit and attempt to win tricks by following priority rules. Once the talon is exhausted, perfect information is enabled via card tracking. Deal winners earn 1-3 game points based on the opponent’s performance, and the first to 7 wins the match.

\subsection{Schnapsen Game Engine}
The implementation of Schnapsen in this paper uses an engine developed by four contributors and published on GitHub - see \cite{isc_schnapsen_repo}. In this implementation, the rules have been slightly simplified. A player cannot voluntarily close the talon. As a consequence, the transition from Phase 1 to Phase 2 occurs only when the talon is exhausted. The engine was coded in Python and includes multiple playable bots that use different strategies and while some excel in Phase 2 by exploiting perfect information, others rely on heuristics. The bots directly used in this project will be: RandBot, BullyBot, and RdeepBot. See Table \ref{tab:bots} for a summary of how these bots operate.

\begin{table}[t]
\centering
\caption{Schnapsen bots used in this research directly.}
\label{tab:bots}

\begin{tabularx}{\textwidth}{l L L L}
\toprule
\textbf{Bot Name} &
\textbf{Decision Logic} &
\textbf{State Usage} &
\textbf{Bot Configuration} \\
\midrule
RandBot &
Uses uniform random selection from valid playing moves. &
Completely ignores game state information. &
Random number generator instance used for deterministic behavior. \\

BullyBot &
Uses priority rules for moving: (1) play trump, (2) follow suit, (3) play highest rank. &
Reactive: uses the current hand but lacks any lookahead. &
Random number generator instance used to break ties between equal-priority moves. \\

RdeepBot &
For each move, generates random valid distributions of hidden cards, simulates play using random policies, and selects the move with the highest average score. &
Handles partial observability by converting unknown game states into concrete sampled states. &
Number of simulations per move (\texttt{num\_samples}) and simulation depth (\texttt{depth}). \\
\bottomrule
\end{tabularx}
\end{table}

Additionally, the MLPlayingBot will be used as a framework for a bot that we will create for the first experiment. MLPlayingBot uses classification to predict the win probability of each valid move using a trained model, based on either a multilayer perceptron architecture or logistic regression. The model is trained on replay memory, which is generated by recording simulated games between two competing bots (e.g., RandBot vs. RandBot). To make decisions, the bot processes information using feature vectors derived from this historical gameplay data.

\subsection{The Schnapsen Challenge}
While the rules of Schnapsen are compact, playing it at a higher level represents a distinct challenge for artificial intelligence. Unlike Chess or Go where the board state is fully observable, Schnapsen players must play without knowing the opponent’s hand or the talon’s order. RdeepBot, the standard baseline used as a benchmark in this paper, represents a relatively strong approach to this problem of (dynamic) partial observability. It generates many hypothetical fully visible game states where the unknown cards are randomly assigned, solves each one using a search algorithm, and averages the results to select the optimal move based on these simulations. However, this strength comes with a cost: it is computationally expensive and relies on fixed search horizons that potentially miss the long-term strategic landscape.

Our vision is to challenge this search-based approach by developing a neural network agent that matches RdeepBot's strategic depth while remaining computationally efficient. In the extreme, a computationally efficient multilayer perceptron would only have one hidden layer. Such neural networks are labeled shallow as opposed to deep \cite{kurkova}. This presents a significant question related to architecture of such candidate: can a shallow multilayer perceptron capture the complex strategy of Schnapsen or does it require a more sophisticated training pipeline? It has been observed that a shallow network may simply not generalize well enough without a sufficiently great amount of training instances or computational capacity \cite{bengio_delalleau}. To investigate this, we first test the limits of supervised learning with a standard shallow network. Subsequently, we explore whether an asynchronous reinforcement learning architecture can successfully generate strategies that lead it to decisively outperform RdeepBot, the search-based baseline. As shown in Table~\ref{tab:bots}, RdeepBot has two important parameters that can be configured - \texttt{num\_samples} and \texttt{depth}. Therefore, we will configure these two parameters to produce variants of RdeepBot, which will influence both the custom bots in the experiments and RdeepBot when acting as a standalone comparison.

\section{Research Questions and Hypotheses}

\subsection{Research Questions}
Based on the focal challenge introduced in the previous section, we define the following research questions:
\\
\\
\textbf{Research Question 1}: \textit{Can an agent using a shallow neural network architecture with supervised learning (MLPBot), trained on replay data generated by RdeepBot variants with different search depths, achieve a statistically significant higher winning rate in Schnapsen than the evaluated baseline RdeepBot depth variants?}
\\
\\
\textbf{Sub-Question 1.1}: \textit{Does the winning rate of MLPBot get better or worse as the search depth parameter of the evaluated RdeepBot opponent increases?}
\\
\\
\textbf{Research Question 2}: \textit{Can an agent using an asynchronous shallow neural network architecture with reinforcement learning (RLBot), trained for 1.2 million games against RdeepBot opponents with different search depths, achieve a statistically significant higher winning rate than the evaluated RdeepBot depth variants?}
\\
\\
\textbf{Sub-Question 2.1}: \textit{Does the winning rate of RLBot get better or worse as the search depth parameter of the evaluated RdeepBot opponent increases?}
\\
\\
\textbf{Research Question 3}: \textit{Can an agent using an asynchronous shallow neural network architecture with reinforcement learning (RLBot), trained for 1.2 million games against RdeepBot opponents with different \texttt{num\_samples} settings, achieve a statistically significant higher winning rate than the evaluated RdeepBot sample variants?}
\\
\\
\textbf{Sub-Question 3.1}: \textit{Does the winning rate of RLBot get better or worse as the \texttt{num\_samples} parameter of the evaluated RdeepBot opponent increases?}

\subsection{Hypotheses}
We propose the following hypotheses corresponding to our research questions:
\\
\\
\textbf{Hypothesis 1}: \textit{The supervised MLPBot, trained on replay data generated by RdeepBot variants with different search depths, will not achieve a statistically significant higher winning rate than the evaluated baseline RdeepBot depth variants.}

\textit{Rationale}: As discussed in Section 2.3, shallow neural networks may struggle to generalize across the complex strategic landscape of Schnapsen without sufficient representational capacity or adaptive learning during interaction. Since MLPBot learns from a fixed replay dataset rather than from direct gameplay feedback, RdeepBot’s explicit lookahead search is expected to retain an advantage in tactical precision, especially in late game situations where more exact calculation matters.
\\
\\
\textbf{Hypothesis 1.1}: \textit{The winning rate of the supervised MLPBot will decrease as the search depth parameter of the evaluated RdeepBot opponent increases.}

\textit{Rationale}: MLPBot relies on a generalized policy learned from static training data, so it is expected to struggle more as the opponent’s search becomes deeper and more precise. As RdeepBot’s lookahead horizon expands, it should exploit weaknesses in MLPBot’s decision-making better, leading to a lower winning rate for MLPBot.
\\
\\
\textbf{Hypothesis 2}: \textit{The asynchronous reinforcement learning agent RLBot, trained for 1.2 million games against RdeepBot opponents with different search depths, will achieve a statistically significant higher winning rate than the evaluated RdeepBot depth variants.}

\textit{Rationale}: Unlike MLPBot, RLBot learns through gameplay interaction and updates its action-value estimates from Monte Carlo outcomes collected in replay memory, rather than merely reproducing patterns from a fixed historical dataset. This training process is expected to allow RLBot to discover strategies that generalize beyond imitation and to exploit weaknesses in RdeepBot’s heuristic search behavior.
\\
\\
\textbf{Hypothesis 2.1}: \textit{The winning rate of RLBot will decrease as the search depth parameter of the evaluated RdeepBot opponent increases.}

\textit{Rationale}: As RdeepBot’s depth increases, its search-based decisions should become stronger and more tactically precise. Therefore, even if RLBot remains competitive or superior overall, defeating deeper RdeepBot variants is expected to become more difficult, reducing RLBot’s winning rate.
\\
\\
\textbf{Hypothesis 3}: \textit{The reinforcement learning agent RLBot, trained for 1.2 million games against RdeepBot opponents with different \texttt{num\_samples} settings, will achieve a statistically significant higher winning rate than the evaluated baseline RdeepBot sample-based variants.}

\textit{Rationale}: RLBot learns a value-based decision policy through large-scale interaction, while RdeepBot’s \texttt{num\_samples} parameter controls how extensively it samples hidden-information states during search. Training RLBot against RdeepBot variants with different sampling strengths is expected to produce a learned agent capable of outperforming the evaluated RdeepBot sample-based variants.
\\
\\
\textbf{Hypothesis 3.1}: \textit{The winning rate of RLBot will decrease as the \texttt{num\_samples} parameter of the evaluated RdeepBot opponent increases.}

\textit{Rationale}: Increasing \texttt{num\_samples} strengthens RdeepBot’s handling of partial observability by improving its approximation over hidden-information states during move selection. As a result, higher-sample RdeepBot variants are expected to be more resilient opponents, leading to a lower winning rate for RLBot.

\section{Experimental Setup}
For a practical implementation of the experiments, we used Python 3.11.11 with built-in and imported modules. The files and source code that we refer to can be requested by the reader via our email addresses mentioned at the beginning of the paper.

\subsection{Note on Asynchronous Processing and Reproducibility}
To ensure that our experiments would be both efficient and reproducible, we utilized asynchronous CPU processing on an Apple M2 Pro chip (10 cores, with 6 for performance and 4 for efficiency) with 16GB of unified memory. We designed our multi-core processing logic using the Python concurrent.futures module \cite{python_concurrent_futures}. In our architecture, we employed “workers” (ProcessPoolExecutor) where each worker roughly maps to one CPU core and typically utilized 8 to 9 cores to retain system stability.

Because the simulation for each experiment was distributed across isolated processes, relying on a global random seed would have been insufficient. Potentially, such an approach could have led to identical game duplications. Therefore, we instead implemented a deterministic seed distribution strategy where we passed a unique “seed\_start” integer to each processing chunk. Inside the worker, the seed for every individual game was calculated linearly as $seed = \text{seed\_start} + i$. This approach ensured that every game played across the many parallel matches had a distinct shuffle and setup, while guaranteeing that the entire experiment could be reproduced by re-running the script with the same initial parameters. This logic can be seen across all provided files but one - the only exception was the 20,000 RdeepBot against RdeepBot game replay memory used to train MLPBot, where the seeds are fixed for both RdeepBot instances.

\subsection{Determining the Suitability of RdeepBot as a Baseline While Manipulating \texttt{depth} and Keeping \texttt{num\_samples} Constant}
Before presenting the setup of the experiments, we will show a result that is a key starting prerequisite: the proof of the strongest available baseline. In Table~\ref{tab:bots} (section 2.2),  we presented three bots from the Schnapsen engine – RandBot, BullyBot, and RdeepBot. To find out which one of these bots has the highest winning rate against the other two opponents, we first determined that we will use three RdeepBot variants based on different search depths: depth 2, depth 4, and depth 6. The number of samples parameter will be fixed at 4. Next, we ran a tournament where each bot played 10,000 games against every other bot. To confirm superiority, we post-processed the resulting winning rate matrix using a one-sample Z-test. For each matchup, we tested the null hypothesis that the winning rate is equal to 50\% against the alternative hypothesis that it differs in a significant way. Given the sample size of $n=10,000$, the standard error was calculated as $SE = \sqrt{\frac{0.5 \times 0.5}{10,000}}$. We categorized the outcomes based on a two-tailed p-value with a significance level of $\alpha = 0.05$:
\begin{itemize}
\item \textbf{B (Better):} Win rate $> 0.5$ and $p < 0.05$.
\item \textbf{W (Worse):} Win rate $< 0.5$ and $p < 0.05$.
\item \textbf{E (Equal):} No statistically significant deviation from 0.5 ($p \ge 0.05$).
\end{itemize}

\begin{center}
  \includegraphics[width=0.7\textwidth]{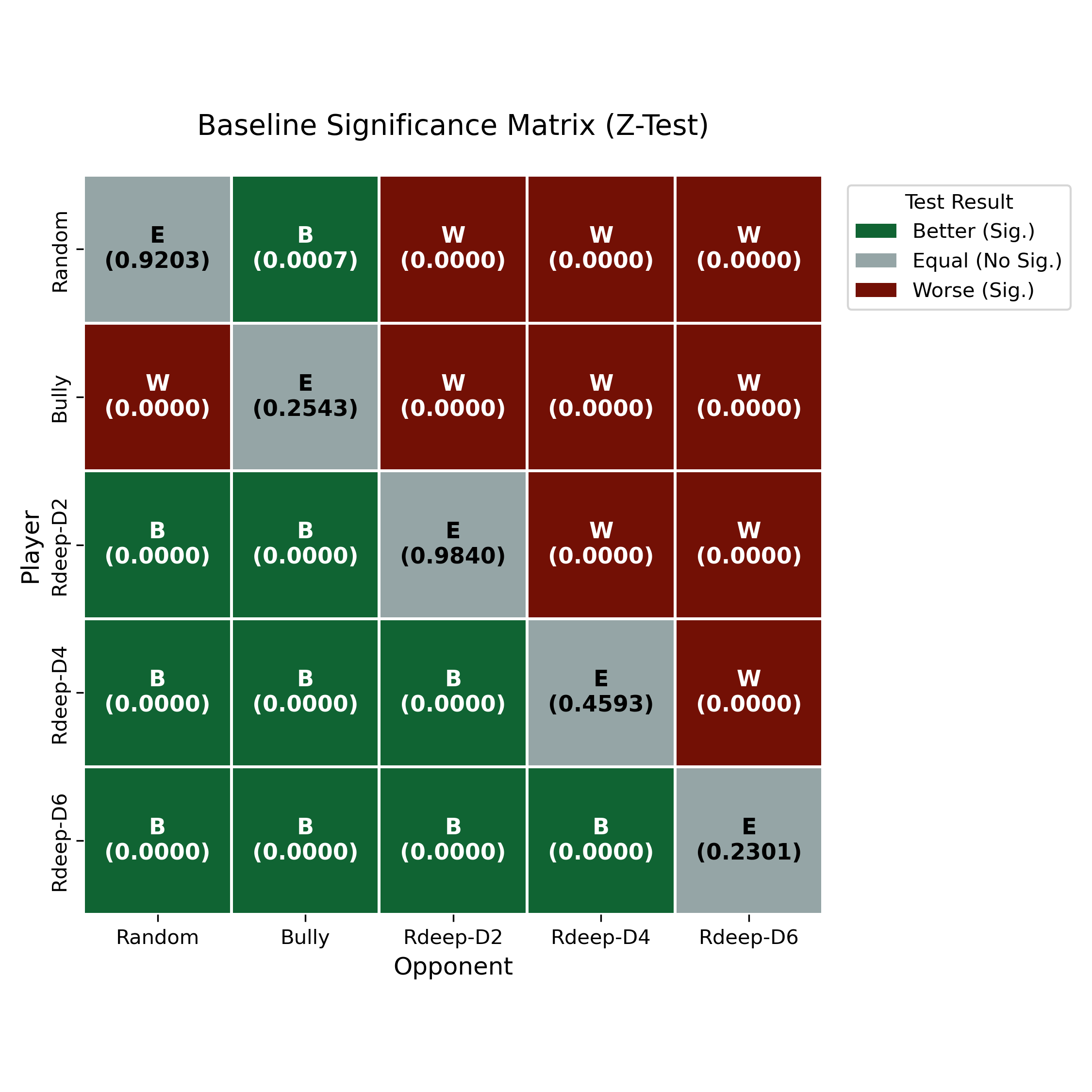}
  \captionof{figure}{Statistical significance matrix displaying Z-Test scores calculated on the winning rates of 10,000 matches between each pair of bots, and the evaluation B(etter), W(orse), and E(qual) based on the logic from Section 4.2.}
  \label{fig1}
\end{center}

As shown in Figure~\ref{fig1}, RdeepBot (no matter whether the depth is 2, 4 or 6) is achieving a higher winning rate than RandBot and BullyBot on a 5\% significance level. Therefore, based on the \texttt{depth} only, RdeepBot is stronger than RandBot and BullyBot, thus providing a candidate baseline. To be certain that RdeepBot can also be a baseline for the Hypotheses 3 and 3.1 that are focused on RdeepBot's \texttt{num\_samples} parameter, we will further investigate in the following section.

\subsection{Determining the Effect of Increasing the RdeepBot's \texttt{num\_samples} Parameter}
An additional behavior that we investigate before presenting the experiments is the effect of increasing the number of samples of RdeepBot on its performance. This is useful prior knowledge for Experiment 3, which will address Hypotheses 3 and 3.1. We define a player RdeepBot with \texttt{num\_samples} set to 4, which is the value that we set for all the RdeepBot depth-based variants in the previous section. This RdeepBot-S4 will be a baseline player, competing against its equivalent S4 variant as well as RdeepBot variants with higher number of samples (20, 40, 60, and 80). For robustness, we perform this test on three cases determined by different search depths (\texttt{depth} of 4, 6, and 8). The results are shown in Figure~\ref{fig2}.

\begin{center}
  \includegraphics[width=0.8\textwidth]{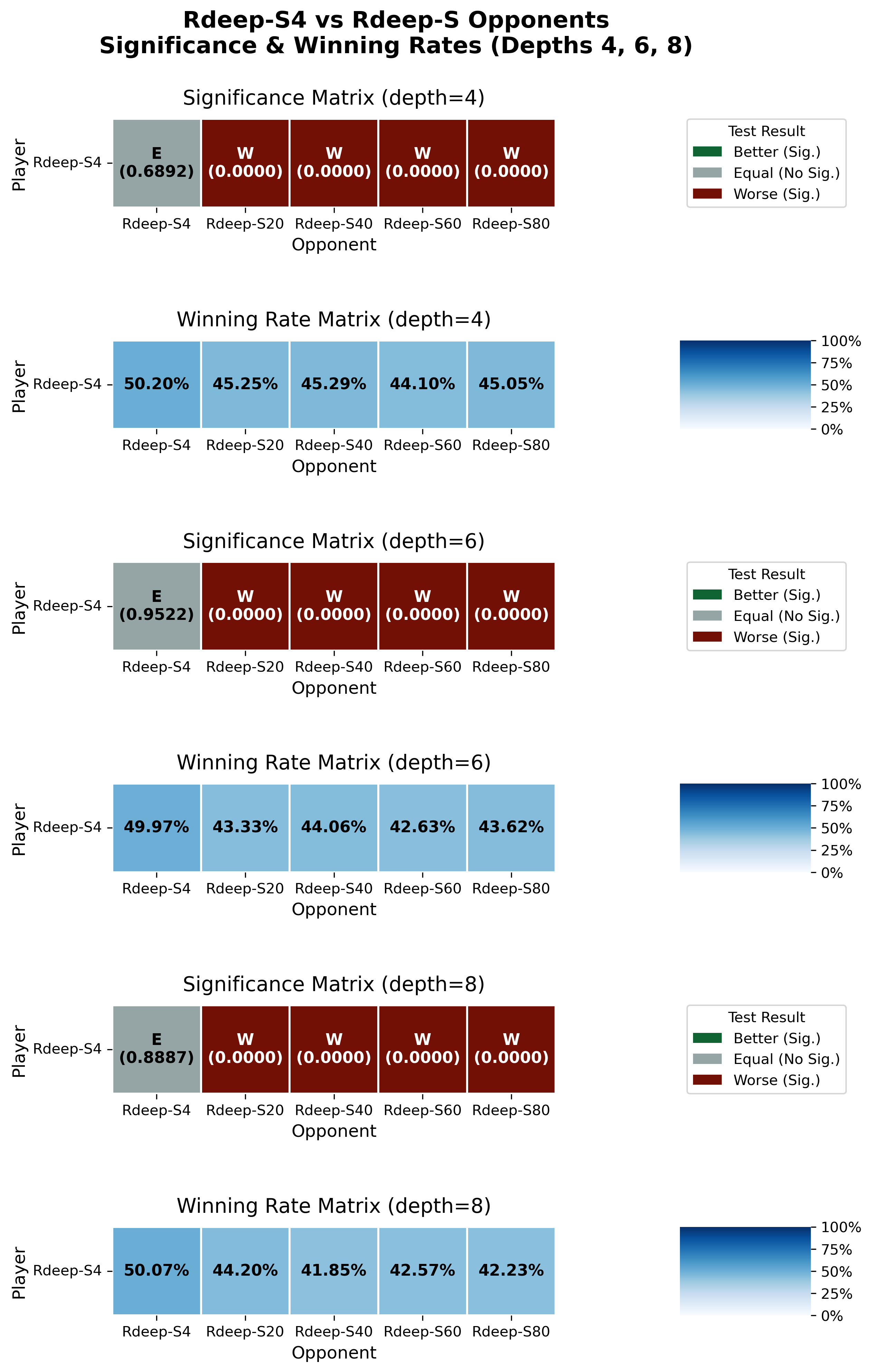}
  \captionof{figure}{Performance of RdeepBot sample variants: Statistical significance matrix displaying Z-Test scores calculated on the winning rates between RdeepBot variants with different \texttt{num\_samples} across search depths of 4, 6, and 8.}
  \label{fig2}
\end{center}

According to the results, Rdeep-S4 against Rdeep-S4 provided no significant winning rate value - this makes sense intuitively, since playing against itself should be too varying to provide any statistically significant information. However, the other variants with higher number of samples than Rdeep-S4 all outperformed the baseline at a statistically significant level. This was consistent for all three depths. This shows that when increasing the number of samples from 4 to a higher amount, there is a clear advantage in the performance of RdeepBot's algorithm. However, as number of samples increases from 20 to 40, 40 to 60, and 60 to 80, this trend is not consistent. Therefore, likely, as number of samples increases even further, it does not guarantee an increasing advantage in performance. Due to this, we will not investigate larger number of samples than 80 in Experiment 3. Importantly, now we can make a certain statement about the performance of RdeepBot compared to RandBot and BullyBot. Whether considering \texttt{depth} or \texttt{num\_samples}, RdeepBot is the strongest available option out of the three types of bots - hence a suitable baseline for our study.

\subsection{Experiment 1: MLPBot Against RdeepBot}
The first experiment explores the supervised architecture against the search-based architecture of RdeepBot. We created an MLPBot (Multilayer Perceptron Bot), which is inspired by the MLPlayingBot from the Schnapsen engine. However, instead of using scikit-learn, we programmed the bot in PyTorch. MLPBot has an input layer, one hidden layer (512 neurons), one activation layer (ReLU), and one output layer (Sigmoid). It uses supervised learning (namely classification) and learns from 20,000 games between RdeepBot (seed 4564654644) and RdeepBot (68438). There are three versions of MLPBot based on three different training samples, each sample based on different search depth of the two RdeepBot competitors (depths 2, 4, and 6, respectively).

The input of the network is a feature vector of size 173, which represents the state of the game combined with a specific move. For specific information included in the feature vector, the reader can refer to the Schnapsen Game Engine (ml\_bot.py file in the source code) \cite{isc_schnapsen_repo}. Our model was trained for 100 epochs using Adam optimizer with a learning rate of $5 \times 10^{-4}$ and a weight decay of $1 \times 10^{-5}$.  The optimizer Adam was chosen due to its proven robustness, memory efficiency, and adaptive learning rate capabilities – overall, it is offering faster convergence compared to other standard stochastic gradient descent methods \cite{adam_optim}. We utilize Binary Cross Entropy (BCELoss) as the loss function. For binary target variables (in our case, 0 representing game loss and 1 representing game victory), the cross-entropy error function is defined as the negative log-likelihood of the likelihood function \cite{bishop}. Consequently, minimizing this loss is mathematically equivalent to maximizing the likelihood of the model’s prediction given the training data \cite{bishop}.

During gameplay, the MLPBot does not perform a tree search. Instead, it evaluates all current valid moves by passing their feature vectors through the network to ultimately predict a winning probability. Afterwards, the bot greedily selects the move with the highest probability. To evaluate performance, we conducted a tournament in which each MLPBot variant played 10,000 games against the RdeepBot baselines.

\subsection{Experiment 2: RLBot-D against RdeepBot-D}
In the second experiment, we investigate a reinforcement learning (RL) approach. Here, the agent learns dynamically through gameplay interaction rather than using static supervised learning. We developed an RLBot (Reinforcement Learning Bot), which has the same shallow neural network architecture as the MLPBot: input layer (173 features), hidden layer (512 neurons) with ReLU activation, and a Sigmoid output layer. However, we used a fundamentally different training methodology.

We approximate the action-value function $Q_{\pi}(s,a)$ as defined by Sutton and Barto \cite{sutton} with a neural network $Q_{\theta}(s,a)\in[0,1]$, interpreted as the predicted probability of winning the Schnapsen deal after taking action $a$ in state $s$.

Specifically, at the completion of each game, we store the episode's sequence of state-action feature vectors $x_t=\phi(s_t,a_t)\in\mathbb{R}^{173}$ in a replay buffer, together with the final binary outcome $G\in\{0,1\}$ (loss/win). 
The central training process samples mini-batches from the replay buffer and
updates the network parameters $\theta$ to minimize the mean squared error
between the predicted action value $Q_\theta(s_t,a_t)$ and the Monte Carlo
outcome target $G$:
\[
\mathcal{L}(\theta)=\mathbb{E}\bigl[(Q_\theta(s_t,a_t)-G)^2\bigr].
\]
During training we use an $\epsilon$-greedy behavior policy \cite{sutton} with $\epsilon$ linearly decayed from $0.23$ to $0.02$ over $1.2$ million games to encourage exploration early in training while gradually shifting toward greedy action selection as the learned action-value function stabilizes.

We trained four separate versions of the RLBot by playing against RdeepBot opponents with search depths of 2, 4, 6, and 8, respectively. The number of samples parameter (\texttt{num\_samples}) of RdeepBot was always set to 4. Finally, we will obtain the following RLBot variants: RL-D2, RL-D4, RL-D6, and RL-D8.

\subsection{Experiment 3: RLBot-S Against RdeepBot-S}
In Experiment 3, we will investigate the performance of RLBot against the baseline RdeepBot (and also show the performance against itself) as we did in Experiment 2 - however, this time we will define the RdeepBot variants with changing \texttt{num\_samples} instead of changing \texttt{depth}. The search depth will be fixed at value 4, while number of samples will be informed by the findings from section 4.3, giving \texttt{num\_samples} of 4, 20, 40, 60, and 80. RLBot will be trained by playing against these RdeepBot variants over 1.2 million games (same as in Experiment 2), yielding the following RLBot variants: RL-S4, RL-S20, RL-S40, RL-S60, RL-S80. The learning logic during training remains unchanged from Experiment 2.

\subsection{Statistical Evaluation Method}
To evaluate the performance of bots in the experiments, we conducted tournaments. Every bot configuration played a set of $n=10,000$ games against every opponent. The raw winning rates were used to compute a one-sample Z-test for each tournament to determine statistical significance.
For each match, we tested the null hypothesis $H_0: p = 0.5$ (assuming equal bot skill) against the alternative hypothesis $H_1: p \neq 0.5$ (indicating a significant skill difference). Given the large sample size, we approximated the binomial distribution with a normal distribution. The standard error (SE) under the null hypothesis was calculated as:
\[
SE = \sqrt{\frac{0.5 \times 0.5}{n}} = \sqrt{\frac{0.25}{10,000}} = 0.005
\]
We computed the Z-score for each winning rate observation $w$ as $Z = \frac{\hat{p} - 0.5}{SE}$, and determined corresponding two-tailed p-values. Using a significance level of $\alpha = 0.05$, we categorized the outcomes into three classes:
\begin{itemize}
\item \textbf{B (Better):} The bot significantly outperformed its opponent ($w > 0.5$ and $p\text{-value} < 0.05$).
\item \textbf{W (Worse):} The bot significantly underperformed its opponent ($w < 0.5$ and $p\text{-value} < 0.05$).
\item \textbf{E (Equal):} There was no statistically significant difference in performance ($p\text{-value} \geq 0.05$), suggesting the bots have comparable gameplay strength.
\end{itemize}
This classification ensures that small deviations from a winning rate of 50\% are not readily interpreted as gameplay superiority unless they are statistically robust.

\subsection{Mechanism and Contribution of RLBot's Replay Buffer}
As part of the training code for RLBot in experiments 2 and 3, we use a replay buffer. It is a finite memory structure in the form of a Python list, which stores past agent interactions. It allows the algorithm to randomly sample mini-batches of historical data for training, instead of solely relying on consecutive on-policy transitions. We set the maximum capacity of the buffer to 100,000 entries, and it acts as a First-In-First-Out queue. As worker processes generate the game trajectories, tuples containing the state (as a vector) and the reward (as a float 1.0 in case of victory, else 0.0) related to the state are pushed to a shared queue and appended to the central buffer. When optimizing, the model uniformly samples random mini-batches of 1,024 states to compute the loss between the network’s predictions and the actual game outcomes.

The idea behind using such a replay buffer for a reinforcement learning agent is based on the mechanism of experience replay \cite{lin1992self}. This concept describes a memory mechanism, due to which the agent remembers its past interactions and presents them to its learning algorithm. In our case, we implement the memory as a sampling mechanism. Our algorithm samples into the buffer uniformly, and while this is not always an optimal approach \cite{krutsylo2025nonuniformmemorysamplingexperience} we proceed with it. We recognize this as a potential limitation, but do not tackle it in our study due to the limited scope of our research.

\section{Results}

\subsection{Results From Experiment 1}
\begin{figure}[H]
  \centering

  \begin{subcaptionblock}{\textwidth}
    \centering
    \includegraphics[width=0.90\linewidth]{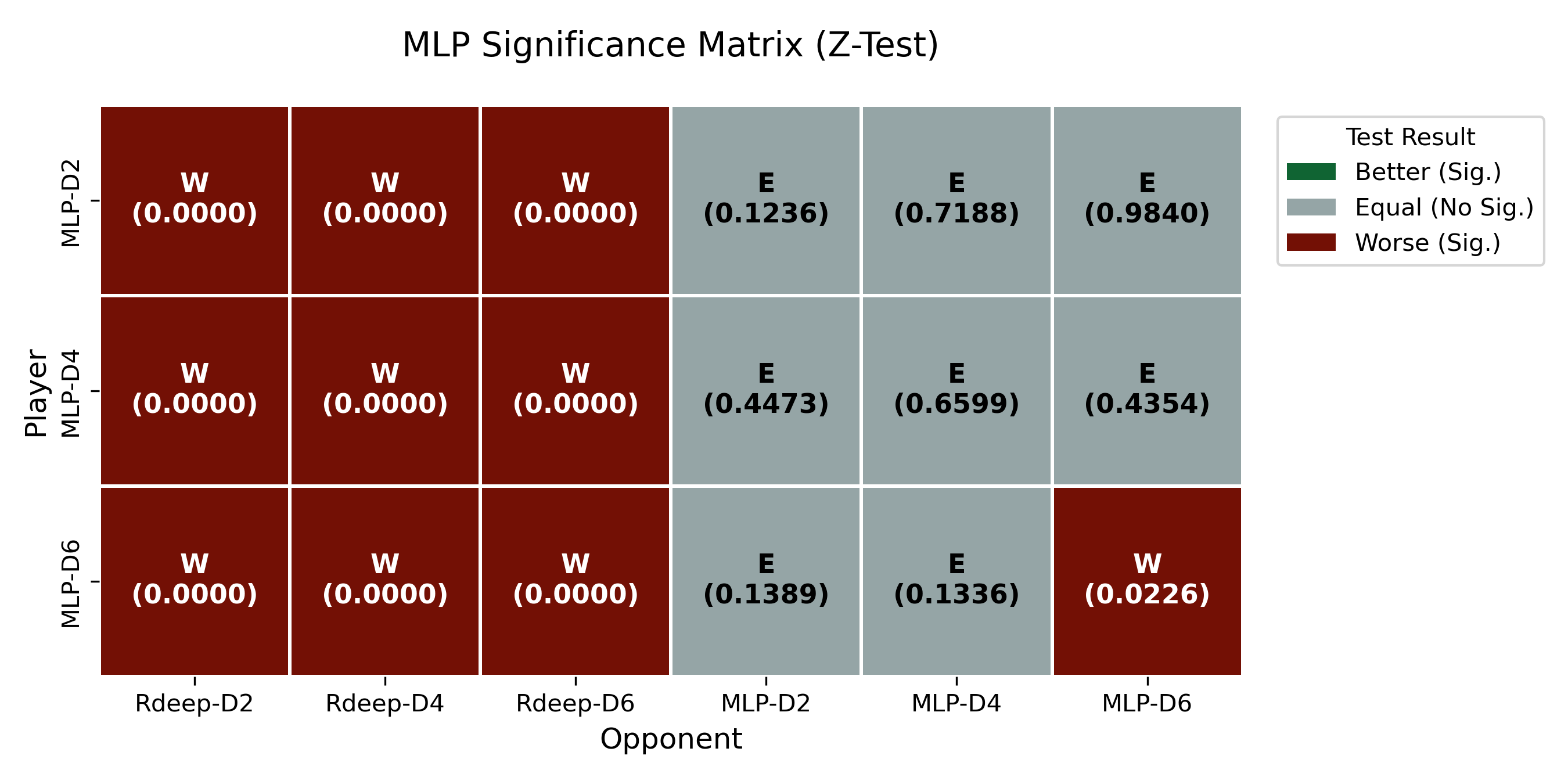}
    \caption{Statistical Significance (Z-Test)}
    \label{fig:mlp_sig}
  \end{subcaptionblock}

  \medskip

  \begin{subcaptionblock}{\textwidth}
    \centering
    \includegraphics[width=0.90\linewidth]{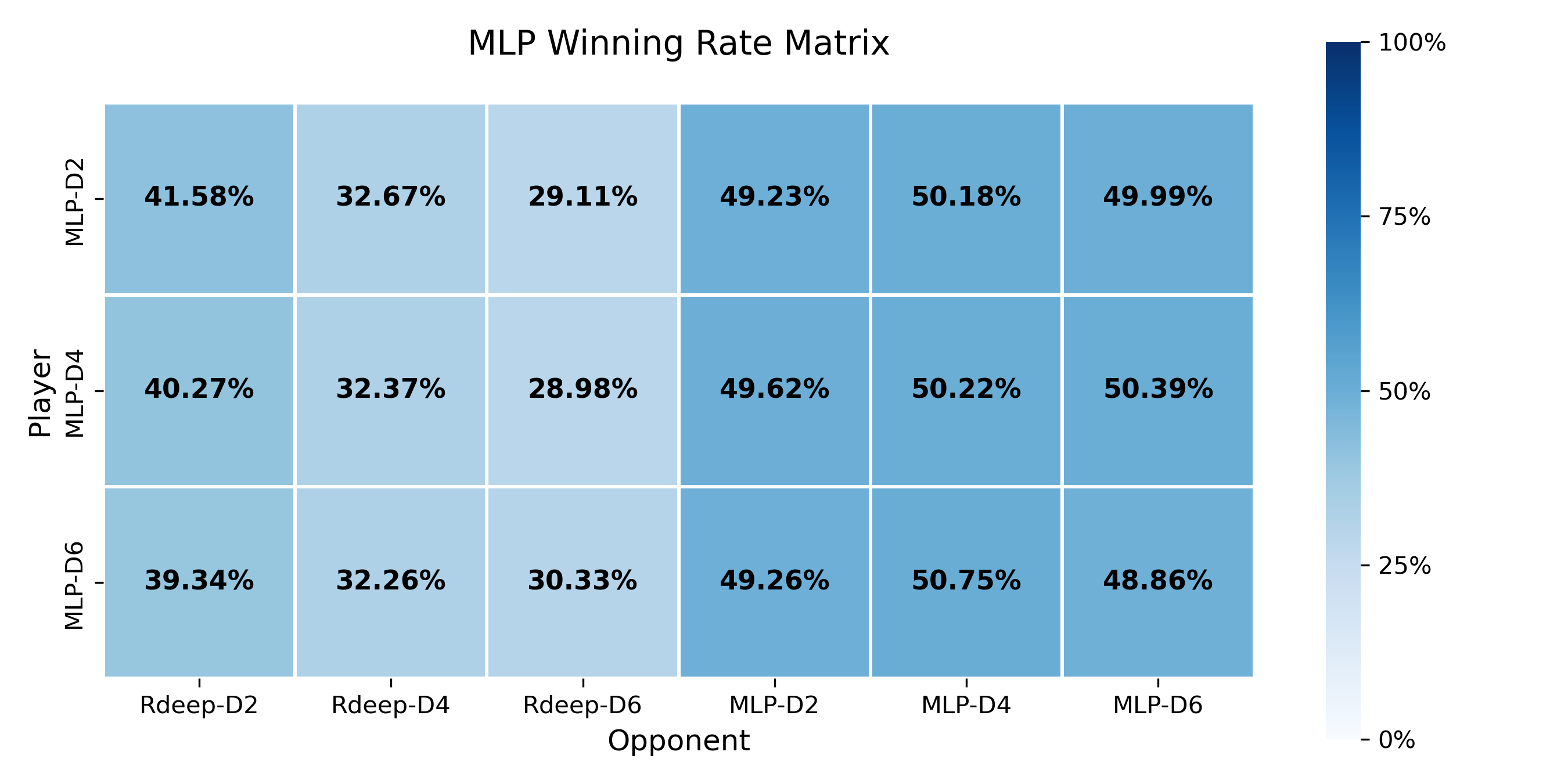}
    \caption{Winning Rates}
    \label{fig:mlp_win}
  \end{subcaptionblock}

  \caption{Experiment 1 Results: Performance of supervised MLPBot variants. Top: Statistical significance categories (Better/Equal/Worse) with p-values. Bottom: Winning rates percentages.}
  \label{fig3}
\end{figure}

The results of the supervised learning experiment are presented in Figure~\ref{fig3}. The top panel displays the statistical significance classifications (Better/Worse/Equal), while the bottom panel shows the winning rates for each opponents' match.

All MLPBot variants were significantly outperformed by the RdeepBot baselines. In every match between an MLPBot and RdeepBot, the Z-test provided a p-value of approximately $0.0000$ and the winning rate of MLPBot was below 50\%. As a consequence, all these matches were classified as "Worse" (W). The winning rates for these games ranged from a minimum of 28.98\% (MLP-D4 versus Rdeep-D6) to a maximum of 41.58\% (MLP-D2 versus Rdeep-D2).

In the matches between the MLPBot variants themselves, the majority of results were classified as "Equal" (E). For instance, MLP-D2 achieved a 50.18\% winning rate against MLP-D4 ($p=0.7188$) and 49.99\% against MLP-D6 ($p=0.9840$), indicating no statistically significant difference in performance. The only statistically significant deviation among these cases was the match between MLP-D6 and MLP-D6, where the Player MLPBot slightly underperformed itself. Overall, increasing the search depth of the teacher RdeepBot instances from 2 to 6 for training data generation did not result in a statistically significant increase in winning rate against the baselines.

\subsection{Results From Experiment 2}
\begin{figure}[H]
  \centering

  \begin{subcaptionblock}{\textwidth}
    \centering
    \includegraphics[width=0.9\linewidth]{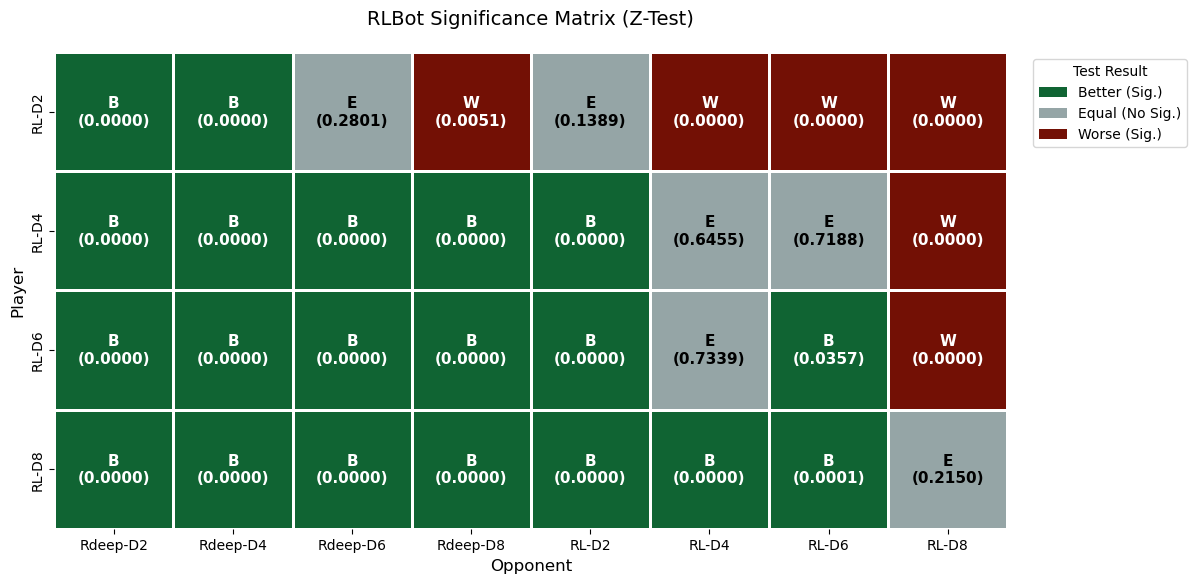}
    \caption{Statistical Significance (Z-Test)}
    \label{fig:rl_d_sig}
  \end{subcaptionblock}

  \medskip

  \begin{subcaptionblock}{\textwidth}
    \centering
    \includegraphics[width=0.9\linewidth]{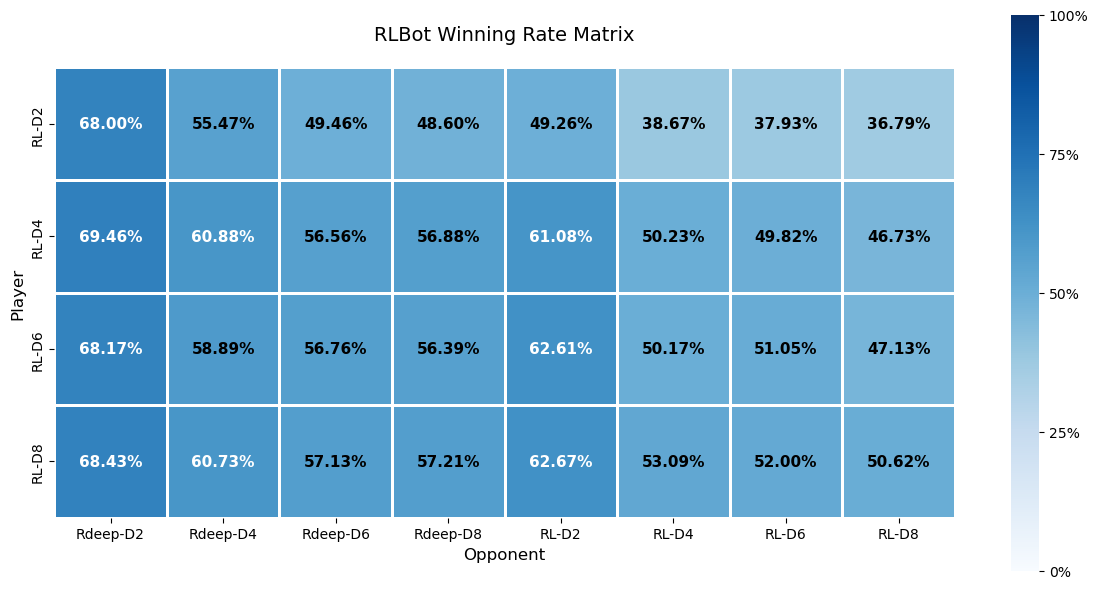}
    \caption{Winning Rates}
    \label{fig:rl_d_win}
  \end{subcaptionblock}

  \caption{Experiment 2 Results: Performance of depth-based RLBot variants. Top: Statistical significance categories (Better/Equal/Worse) with p-values. Bottom: Winning rates percentages.}
  \label{fig4}
\end{figure}

The results of Experiment 2 are presented in Figure~\ref{fig4}. The RLBot agents demonstrated a strong ability to generalize against the RdeepBot opponents. Unlike the supervised MLPBot, the majority of variants of RLBots maintained positive winning rates against the strongest baselines. For example, the RL-D8 variant achieved statistically significant wins against all RdeepBot versions, securing a 57.21\% win rate against Rdeep-D8 and 60.73\% against Rdeep-D4. Similarly, RL-D4 performed exceptionally well, achieving a 56.88\% win rate against the strongest baseline, Rdeep-D8.

Furthermore, the results demonstrate a positive correlation between search depth and playing strength. While all RL agents share the same neural network weights, their performance varies based on the lookahead horizon available during inference. For instance, RL-D2 proved effective against weaker opponents (68.00\% against Rdeep-D2) but hit a performance ceiling against stronger agents. It was statistically indistinguishable from Rdeep-D6 and performed significantly worse against Rdeep-D8. Additionally, when RLBot agents played against each other, deeper search depths provided a consistent advantage. RL-D8 achieved a statistically significant winning margin against RL-D2, RL-D4, and RL-D6. This trend holds across all variants, where increases in search depth consistently yielded statistically significant improvements in the winning rate.

\subsection{Results From Experiment 3}

\begin{figure}[H]
  \centering

  \begin{subcaptionblock}{\textwidth}
    \centering
    \includegraphics[width=0.8\linewidth]{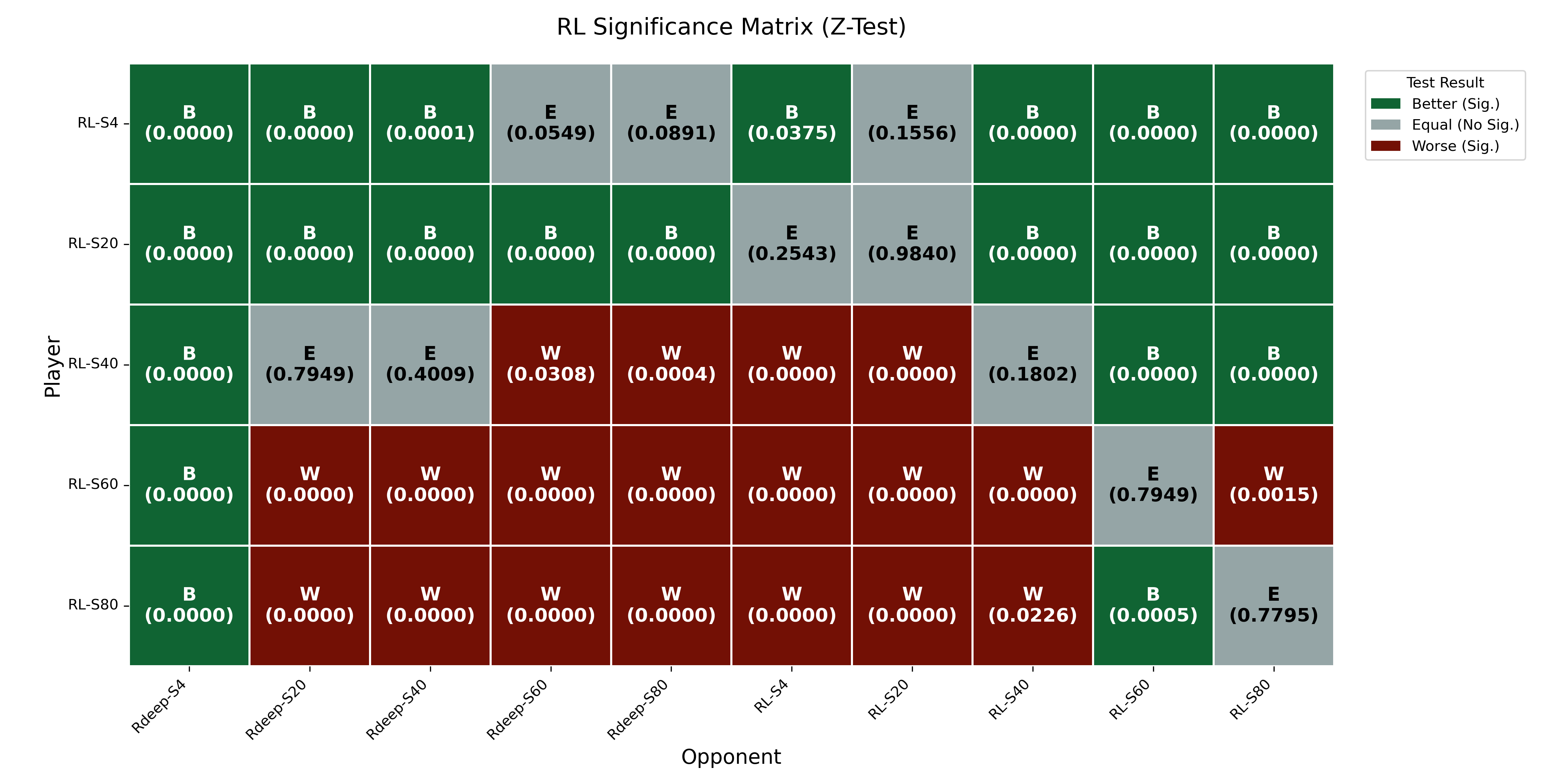}
    \caption{Statistical Significance (Z-Test)}
    \label{fig:rl_s_sig}
  \end{subcaptionblock}

  \medskip

  \begin{subcaptionblock}{\textwidth}
    \centering
    \includegraphics[width=0.8\linewidth]{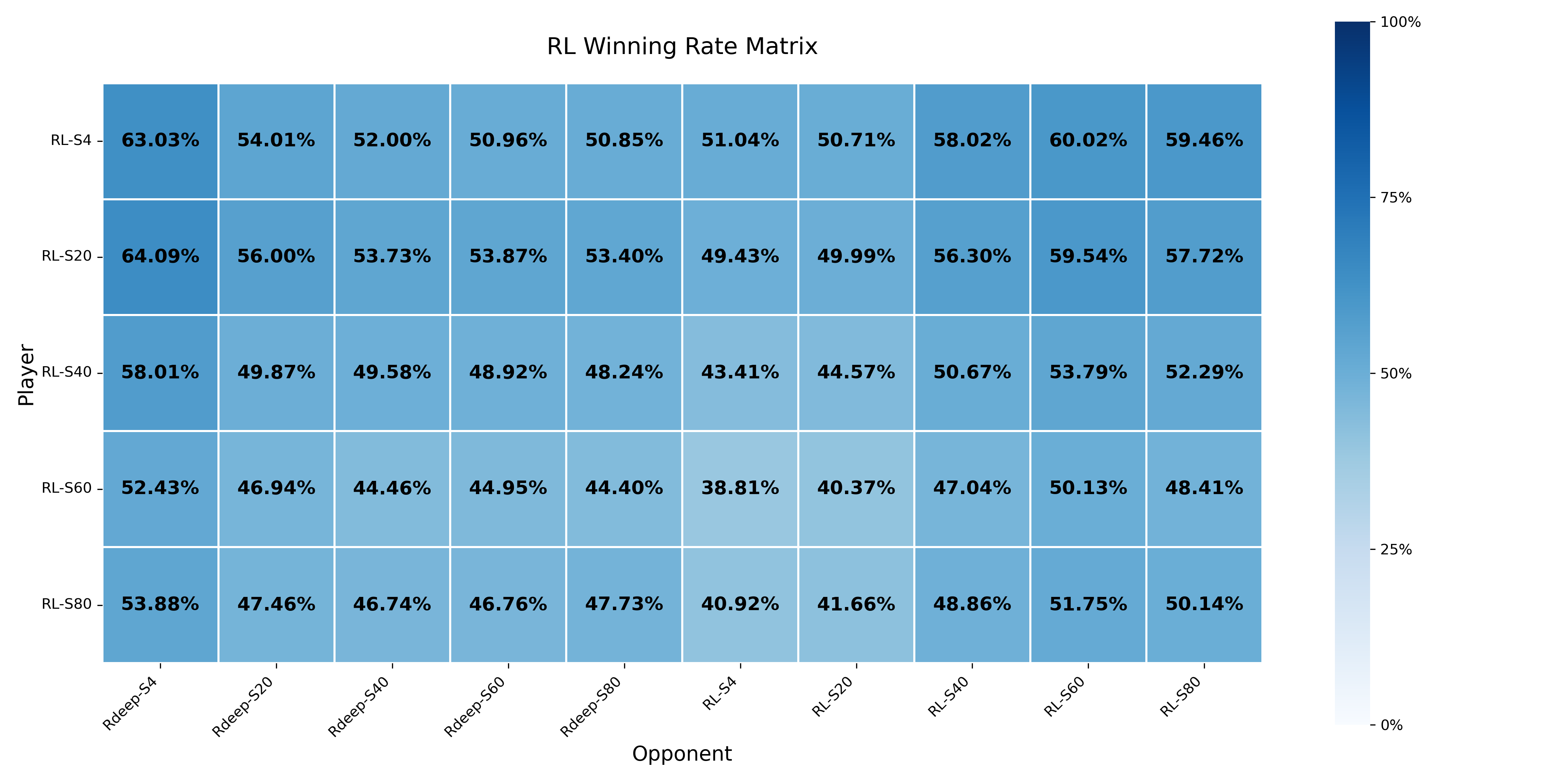}
    \caption{Winning Rates}
    \label{fig:rl_s_win}
  \end{subcaptionblock}

  \caption{Experiment 3 Results: Performance of sample-based RLBot variants. Top: Statistical significance categories (Better/Equal/Worse) with p-values. Bottom: Winning rates percentages.}
  \label{fig5}
\end{figure}

The results from Experiment 3 can be seen in Figure~\ref{fig5}. Notably, RL-S20 achieved a statistically significant better performance over all the RdeepBot-S variants. The winning rate of RL-S20 has a generally decreasing trend as \texttt{num\_samples} of RdeepBot-S opponents increases. This general trend is consistent among all the RLBot variants. However, for RL-S4 and RL-S40, not all of the results are statistically significant (for example, the winning rate of RL-S40 against Rdeep-S20 or the one of RL-S4 against Rdeep-S60). In general, the results suggest that as the number of samples of RdeepBot (used to train RLBot) increase from 4 to 20, the winning rate of RLBot against the RdeepBot opponents gets higher (and, in addition, higher count of statistically significant results is obtained). However, as the number of samples of RdeepBot used to train RLBot increases from \texttt{num\_samples} of 20 to \texttt{num\_samples} of 40 and from \texttt{num\_samples} of 40 to \texttt{num\_samples} of 60, the winning rate of RLBot decreases. RL-S80 achieves a slightly higher winning rates against RdeepBot than RL-S60, however, it is still lower than the winning rates of all the other RLBot variants.

\section{Discussion}
To begin with, in the depth-based Experiment 1, all MLPBot variants were significantly outperformed by the RdeepBot baselines - the winning rates against RdeepBot ranged from 28.98\% to 41.58\%. This supports Hypothesis 1 and suggests that supervised training based on fixed replay data was not sufficient to produce a policy that could outperform a search-based approach.

The results also support Hypothesis 1.1. As the search \texttt{depth} of the RdeepBot opponent increases, MLPBot's performance decreases across all the tested variants. For example, MLP-D4 has a winning rate of 40.27\% against Rdeep-D2 and a lower winning rate of 32.37\% against Rdeep-D4. This is consistent with the intuition that deeper opponent search makes the RdeepBot baseline more difficult for the supervised bot to outperform.

At the same time, the results show an additional pattern concerning the quality of the training data. Increasing the \texttt{depth} of the RdeepBot instances used to generate the replay data for MLPBot did not yield substantial downstream improvement against the baselines. For each opponent, the three MLPBot variants remain relatively close to one another: the spread between the lowest and highest winning rate is 2.24 percentage points against Rdeep-D2, 0.41 percentage points against Rdeep-D4, and 1.35 percentage points against Rdeep-D6. This suggests that simply increasing teacher \texttt{depth} is not sufficient to substantially strengthen the supervised agent within this experimental setup. More cautiously, the pattern is consistent with a limitation of the imitation-based approach itself, which may partly stem from the fact that MLPBot is trained on a fixed replay distribution and then evaluated without online adaptation or search.

The depth-based Experiment 2 provides additional interesting insights. The majority of the 16 matchups between RLBot variants and the RdeepBot opponents were won by RLBot at a statistically significant level. Notably, the winning rate of RL-D4, RL-D6, and RL-D8 was higher than 56\% against the Rdeep-D8, which is the opponent with the highest search \texttt{depth} in our experiment. These results support Hypothesis 2 and suggest that an asynchronous algorithm with Monte Carlo training and a built-in memory buffer enabled the network to learn a much stronger action-value estimate than supervised imitation.

The evidence for Hypothesis 2.1 is more mixed. RL-D2 declined from 68\% against Rdeep-D2 to below 50\% against Rdeep-D6 (not statistically significant) and against Rdeep-D8 (statistically significant). This is consistent with the expectation that stronger opponent search makes it harder for the player bot to outperform on the winning rate. However, the winning rates of the deeper RLBot variants remained superior at a statistically significant level, even for higher RdeepBot opponent depths. For illustration, RL-D6 has a winning rate of 58.89\% against Rdeep-D4 and 56.39\% against Rdeep-D8, which is still a superior level when compared to the RL-D2's decrease from 55.47\% to 48.60\% for the same opponents. This suggests that the predicted downward trend does not hold uniformly once the learned evaluator is paired with a sufficiently deep inference.

Overall, reinforcement learning allowed RLBot to achieve statistically significant wins against strong RdeepBot baselines, showing that the learned \(Q(s,a)\) estimate captured strategically useful information. However, the matches between RLBot variants themselves show that deeper lookahead during move selection still improved performance. In the Results section, we report that RL-D8 significantly outperformed RL-D2, RL-D4, and RL-D6. Therefore, the most appropriate interpretation is not that the network replaces search, but that the learned value estimate becomes more effective when combined with deeper lookahead in gameplay.

The sample-based Experiment 3 provides weaker and more conditional support for Hypothesis 3 than the depth-based Experiment 2 provided for Hypothesis 2. The strongest evidence comes from RL-S20, which achieved a statistically significant better performance against all the evaluated RdeepBot-S variants. However, this pattern does not extend uniformly to all sample-based RLBot variants - as shown in Results, some comparisons were not statistically significant, and performance declined once the \texttt{num\_samples} of the RdeepBot used for RLBot training increased to the cases beyond \texttt{num\_samples} of 20. Therefore, it cannot be concluded that sample-based reinforcement learning consistently outperforms the RdeepBot-S baselines across all settings. Rather, the evidence suggests that this training setup could produce a strong agent, but only when the \texttt{num\_samples} value of the RdeepBot opponent used during training was set within a limited range.

The evidence for Hypothesis 3.1 is also supportive, but not perfectly uniform. RL-S20 shows a generally decreasing trend as \texttt{num\_samples} of the RdeepBot-S opponents increases, and this general pattern is consistent across all the RLBot variants. Thus, the underlying intuition of Hypothesis 3.1 is supported - stronger sampling makes RdeepBot more difficult to defeat because it handles hidden information more effectively. At the same time, not every comparison is statistically significant or perfectly monotonic, due to which the hypothesis is better described as generally supported than as uniformly confirmed in every case.

Together, the three experiments point to a consistent conclusion. Shallow supervised learning was not an efficient approach to defeat the search-based RdeepBot benchmark, whereas shallow reinforcement learning produced agents that generalized much better against strong opponents. At the same time, the best results were not obtained by increasing complexity indiscriminately, whether in the depth-based or the sample-based setting. In the former setting, the best results were obtained by combining a learned value function with sufficiently strong lookahead and, in the latter, by training against an opponent of intermediate rather than maximal difficulty.

\section{Conclusion and Future Works}
In this study, we demonstrated that reinforcement learning provides a substantially more effective framework for mastering Schnapsen than supervised imitation. While MLPBot remained limited by the fixed distribution and quality of its replay-based training data, RLBot learned a stronger value estimate through asynchronous Monte Carlo updates and experience replay. This difference in training paradigm allowed RLBot to generalize more successfully against strong RdeepBot opponents in the evaluation tournaments.

Additionally, the tournament results showed that learning alone was not the full explanation of performance. In the depth-based setting, RLBot was trained against RdeepBot opponents with different search depths, and the resulting trained variants were then evaluated against RdeepBot depth-based baselines in the tournament. The strongest results were obtained when the learned value function was combined with deeper lookahead during gameplay, with the deepest RLBot variant achieving statistically significant higher winning rates against all evaluated RdeepBot depth-based opponents.

At the same time, the sample-based experiments showed that stronger training opponents were not always better for learning. In Experiment 3, RLBot was trained separately against RdeepBot opponents with different \texttt{num\_samples} values, and these trained RLBot variants were then evaluated in tournaments against RdeepBot-S baselines. The results showed that RL-S20 was the strongest sample-based variant, but increasing the \texttt{num\_samples} of the RdeepBot training opponent beyond that point did not continue to improve tournament performance.

Taken together, these findings suggest that shallow reinforcement learning can produce strong Schnapsen agents, but that performance depends both on the training opponent used to shape the learned value function and on the search configuration used during gameplay evaluation. The study shows not only that reinforcement learning is more effective than supervised imitation, but also that the relationship between training difficulty and final playing strength is non-trivial.

For future work, we propose transitioning from training against a fixed heuristic to a full self-play framework in the spirit of AlphaZero. Such an approach could allow the agent to discover novel strategies independent of biases of its opponents. Additionally, investigating architectures such as recurrent neural networks or transformers could better capture card history and partial observability in Schnapsen. Finally, future work could examine improved replay buffer strategies, since the current approach uses uniform sampling despite recognizing it as a potential limitation. Together, these directions could help reduce reliance on explicit search while further improving strategic generalization.

\bibliographystyle{plainurl} 
\bibliography{references} 

\end{document}